\PassOptionsToPackage{table,xcdraw,dvipsnames}{xcolor}

\documentclass{article}

\PassOptionsToPackage{numbers, compress}{natbib}
\usepackage[preprint]{neurips_2024}

\usepackage[utf8]{inputenc} 
\usepackage[T1]{fontenc}    
\usepackage{hyperref}       
\hypersetup{
    colorlinks=true,
    linkcolor=BlueViolet,
    citecolor=teal
}
\usepackage{url}            
\usepackage{booktabs}       
\usepackage{amsfonts}       
\usepackage{nicefrac}       
\usepackage{microtype}      
\usepackage{xcolor}         

\usepackage{makecell}
\usepackage{amsthm}
\usepackage{bm}
\usepackage[table,xcdraw]{xcolor}
\usepackage{multicol}
\usepackage{colortbl}
\usepackage{amsmath}
\usepackage{cleveref}
\usepackage{multirow}
\usepackage{graphicx}
\usepackage{subfig}
\usepackage{enumitem}
\usepackage{bbding}
\usepackage{color}

\usepackage{algorithm}
\usepackage{listings}
\usepackage{natbib}

\usepackage{graphicx}
\usepackage{booktabs}
\usepackage{orcidlink}

\usepackage{dsfont}
\usepackage{multirow}
\usepackage{amsmath}
\usepackage{array}

\usepackage{textcomp}
\usepackage{stfloats}
\usepackage{url}
\usepackage{verbatim}
\usepackage{graphicx}
\usepackage{mathrsfs}

\usepackage{booktabs}
\usepackage{nicefrac}
\usepackage{microtype}
\usepackage{float}
\usepackage{makecell}
\usepackage{pifont}
\usepackage{listings}
\usepackage{enumitem}

\usepackage{color, colortbl}
\usepackage{wrapfig}
\usepackage{changepage,threeparttable}

\usepackage{etoolbox}
\makeatletter
\AfterEndEnvironment{algorithm}{\let\@algcomment\relax}
\AtEndEnvironment{algorithm}{\kern2pt\hrule\relax\vskip3pt\@algcomment}
\let\@algcomment\relax
\newcommand\algcomment[1]{\def\@algcomment{\footnotesize#1}}
\renewcommand\fs@ruled{\def\@fs@cfont{\bfseries}\let\@fs@capt\floatc@ruled
  \def\@fs@pre{\hrule height.8pt depth0pt \kern2pt}%
  \def\@fs@post{}%
  \def\@fs@mid{\kern2pt\hrule\kern2pt}%
  \let\@fs@iftopcapt\iftrue}
\makeatother

\definecolor{ForestGreen}{RGB}{34,139,34}

\newcommand{\ie}{\emph{i.e.},~}
\newcommand{\eg}{\emph{e.g.},~}
\newcommand{\wrt}{\emph{w.r.t.}~}

\renewcommand{\paragraph}[1]{\medskip\noindent\textbf{#1.~}}

\title{GMMFormer v2: An Uncertainty-aware Framework for Partially Relevant Video Retrieval}

\author{Yuting Wang\\
        Tsinghua University\\
        \texttt{wangyt22@mails.tsinghua.edu.cn} \\
        \And
        Jinpeng Wang\\
        Tsinghua University\\
        \texttt{wjp20@mails.tsinghua.edu.cn} \\
        \And
        Bin Chen\thanks{Corresponding author.}\\
        Harbin Institute of Technology, Shenzhen\\
        \texttt{chenbin2021@hit.edu.cn} \\
        \And
        Tao Dai\\
        Shenzhen University\\
        \texttt{daitao@szu.edu.cn} \\
        \And
        Ruisheng Luo\\
        Tsinghua University\\
        \texttt{luors22@mails.tsinghua.edu.cn}
        \And
        Shu-Tao Xia\\
        Tsinghua University\\
        \texttt{xiast@sz.tsinghua.edu.cn} \\
}

\begin{document}

\maketitle

\begin{abstract}
  Given a text query, partially relevant video retrieval (PRVR) aims to retrieve untrimmed videos containing relevant moments. 
  Due to the lack of moment annotations, the uncertainty lying in clip modeling and text-clip correspondence leads to major challenges. 
  Despite the great progress, existing solutions either sacrifice efficiency or efficacy to capture varying and uncertain video moments. 
  What's worse, few methods have paid attention to the text-clip matching pattern under such uncertainty, exposing the risk of \emph{semantic collapse}. 
  To address these issues, we present GMMFormer v2, an uncertainty-aware framework for PRVR. 
  For clip modeling, we improve a strong baseline GMMFormer~\cite{GMMFORMER} with a novel temporal consolidation module upon multi-scale contextual features, which maintains efficiency and improves the perception for varying moments. 
  To achieve uncertainty-aware text-clip matching, we upgrade the query diverse loss in GMMFormer to facilitate fine-grained uniformity and propose a novel optimal matching loss for fine-grained text-clip alignment. 
  Their collaboration alleviates the semantic collapse phenomenon and neatly promotes accurate correspondence between texts and moments. 
  We conduct extensive experiments and ablation studies on three PRVR benchmarks, demonstrating remarkable improvement of GMMFormer v2 compared to the past SOTA competitor and the versatility of uncertainty-aware text-clip matching for PRVR. 
  Code is available at \url{https://github.com/huangmozhi9527/GMMFormer_v2}.
\end{abstract}

\section{Introduction}
\label{sec:intro}

\begin{figure}[t]
  \centering
  \includegraphics[width=\linewidth]{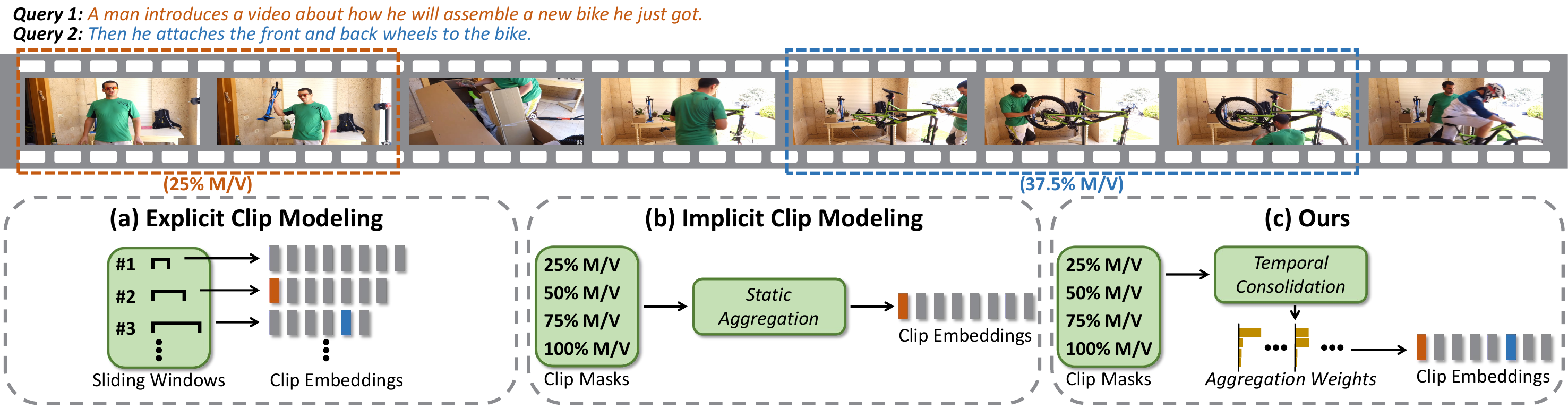}
  \caption{(a) Explicit PRVR methods adopt multi-scale sliding windows to traverse all possible clips, which are redundant and inefficient. 
  (b) Implicit methods improve efficiency by combining multi-scale information and generating fewer clip embeddings. However, the static aggregation is inflexible for capturing moments with unexpected moment-to-video ratios (M/Vs), \eg the clip in the blue dotted box, beyond predefined clip masks. 
  (c) We propose a temporal consolidation module to improve the clip modeling. By learning adaptive aggregation weights for different time points in a video, it is capable of perceiving video moments with varying lengths.
  }
  \label{intro}
\end{figure}

Partially relevant video retrieval (PRVR) aims to find untrimmed videos comprising moments relevant to a given text query.
Recently, PRVR models~\cite{MSSL, PEAN, DKD, GMMFORMER} have become popular as they can better deal with ubiquitous untrimmed videos in practice and reduce human labor of moment annotations.
We identify clip modeling and text-clip matching as two major challenges in PRVR, where the key lies in tackling uncertainty. 
Different from previous probabilistic modeling works~\cite{oh2018modeling, shi2019probabilistic, zhang2021point, chun2021probabilistic, fang2023uatvr}, where uncertainty relates to the noises in data, the uncertainty in PRVR is typically due to the lack of fine-grained annotations, \ie the locations and duration of moments in videos and their associations with the text queries.

Existing PRVR works can be divided into explicit~\cite{MSSL} and implicit~\cite{PEAN, DKD, GMMFORMER} methods regarding clip modeling, which struggle to balance efficiency and efficacy.
As shown in Fig. \ref{intro}(a), explicit models (\eg MS-SL~\cite{MSSL}) often utilize multi-scale sliding window strategies to traverse all possible clips, which suffer from redundancy and inefficiency. 
Implicit models (\eg PEAN~\cite{PEAN} and GMMFormer~\cite{GMMFORMER}) incorporate multi-scale contextual features and generate compact clip embeddings. As illustrated in Fig. \ref{intro}(b), although they improve retrieval efficiency, it is inflexible to capture moments with varying lengths.
Another blocking issue in PRVR is the \emph{semantic collapse}, which refers to a phenomenon where different text queries relevant to a video tend to match a few homogeneous clips. This will result in most informative clip embeddings not receiving supervision signals throughout the training and eventually degrade accurate text-clip matching.

To address the above issues, we propose GMMFormer v2, an uncertainty-aware framework for PRVR. 
For clip modeling, we design a novel temporal consolidation module upon a strong baseline GMMFormer~\cite{GMMFORMER} to improve the clip modeling capacity. Specifically, as shown in Fig. \ref{intro}(c), the model learns aggregation weights for different time points in a video to balance the importance of multi-scale contextual features, enabling flexible perception of video moments in varying lengths. 
To achieve uncertainty-aware text-clip matching, we first upgrade the query diverse loss in GMMFormer by actively focusing on hard samples, which regularizes a uniform and discriminative structure of fine-grained text semantics. 
Besides, we propose an optimal matching loss via the Hungarian algorithm~\cite{kuhn1955hungarian}, guiding different text queries to align to relevant clips within the same video in a fine-grained and diversified manner. 
The collaboration of both losses effectively alleviates the semantic collapse and encourages better text-clip alignment between texts and untrimmed videos.

We conduct extensive experiments on three PRVR benchmarks, including ActivityNet Captions~\cite{activitynet}, TVR~\cite{tvr} and Charades-STA~\cite{sta}. The results demonstrate the remarkable improvement and versatility of our solution. 
In particular, GMMFormer v2 achieves \textbf{6.1\%}, \textbf{7.1\%} and \textbf{7.3\%} relative lifts of SumR compared to GMMFormer on three PRVR benchmarks respectively, and significantly surpasses the past SOTA competitor, DL-DKD~\cite{DKD}, which resorts to extra knowledge of pretrained vision-language models~\cite{CLIP}. In addition, compared to GMMFormer, GMMFormer v2 has better scalability in utilizing varying Gaussian constraints. Finally, uncertainty-aware text-clip matching can be used as \emph{plugin-in} supervision to improve the performance of other PRVR models.

Our main contributions can be summarized as follows:
\begin{itemize}[noitemsep]
    \item We highlight important challenges related to the uncertainty in PRVR: varying moments in videos and unclear fine-grained text-clip correspondence.
    \item We design a novel temporal consolidation module to improve GMMFormer~\cite{GMMFORMER}, which shows better capacity in capturing video moments with varying lengths.
    \item We propose a \emph{plugin-in} solution to alleviate the risk of \emph{semantic collapse} in PRVR, including a revamped query diverse loss for fine-grained uniformity and a novel optimal matching loss to promote fine-grained text-clip alignment.
    \item Extensive experiments and ablation studies on three PRVR benchmarks demonstrate the effectiveness of our proposed framework.
\end{itemize}

\section{Related Work}

\subsection{Partially Relevant Video Retrieval} 

With the development of deep learning~\cite{xie2024unipts, li2024unleashing, wang2023missrec, wang2022tenet, li2023fsr}, 
the explosion of video data has attracted widespread research attention on video analysis tasks~\cite{li2022shaobo, conmh, tg_vqa, magrh, tvts, tvtsv2}. 
Among them, text-to-video retrieval (T2VR)~\cite{jin2022expectation, jin2023video, jin2023text, jin2023diffusionret, wang2022hybrid, liu2022multi, li2023progressive, li2023momentdiff} is an active topic focusing on retrieving short video clips using text queries. 
Typically, T2VR methods assume that each video is pre-trimmed and can be entirely relevant to an oracle text query. However, such requirements are often not met in practice, as real-world videos are often untrimmed and contain a lot of background content~\cite{ji2023binary, qu2020fine, xiao2020visual}. 
Human annotations for informative moments in videos are usually expensive and inefficient. As a result, T2VR models designed for trimmed videos may not adapt well to practical applications.

The task of partially relevant video retrieval (PRVR)~\cite{MSSL} is targeted for practical needs particularly. In PRVR, each video is allowed to match multiple text queries, while each query can be only relevant to one moment in the video. This makes PRVR compatible with practical applications and thus an important task. 
MS-SL~\cite{MSSL} is the first work to define the PRVR task and provides a strong baseline with explicit clip modeling. The generated clip embeddings are redundant and inefficient, consuming a significant storage overhead. 
Recently, several implicit methods~\cite{PEAN, GMMFORMER, DKD} have been proposed to improve the efficiency. 
However, they are still inflexible to capture uncertain moments in untrimmed videos. 
Moreover, to our knowledge, there is little attention paid to the fine-grained semantic structure, given its uncertain nature in PRVR.

\subsection{Uncertainty in Computer Vision}
The uncertainty problem has a long history in computer vision, where most solutions were based on probabilistic embeddings.
For example, HIB~\cite{oh2018modeling} first introduced probabilistic embeddings to capture the uncertainty of image representations while handling the one-to-many correspondence of deep metric learning. 
Similar ideas have been extensively investigated in various tasks, including face recognition \cite{shi2019probabilistic} and instance segmentation \cite{zhang2021point}. \cite{chun2021probabilistic} innovatively applied probabilistic embeddings to cross-modal retrieval, and \cite{fang2023uatvr} further improved uncertainty modeling in T2VR by combining deterministic and probabilistic embeddings. 
Different from them, uncertainty in PRVR is mainly reflected in unknown locations and duration of moments, due to the lack of moment annotations. Given different motivations, we contribute new strategies in model design and training objectives, which are different but orthogonal to those probabilistic solutions. We leave their combination for future work.

\section{GMMFormer v2}

\begin{figure}[t]
  \centering
  \includegraphics[width=\linewidth]{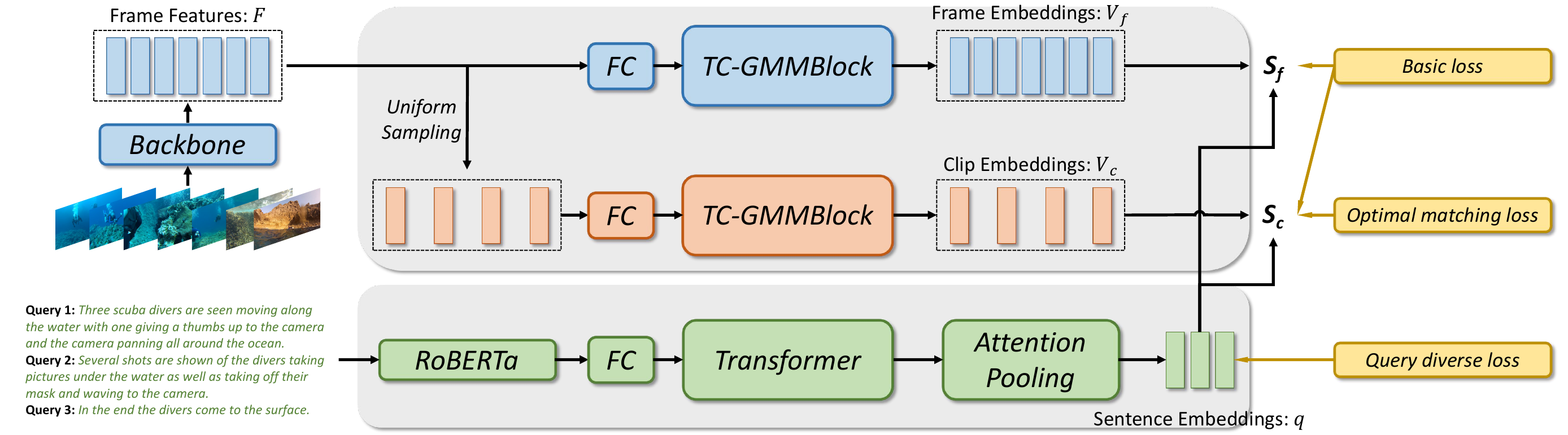}
  \caption{\textbf{The overall architecture of GMMFormer v2.} 
  }
  \label{overview}
\end{figure}

In this section, we describe the uncertainty-aware framework termed GMMFormer v2, including text query representation (Sec. \ref{text_query_representation}), video representation with temporal consolidation (Sec. \ref{video_representation}), uncertainty-aware text-clip matching (Sec. \ref{uncertainty_aware_cross_modal_matching}) and similarity measure (Sec. \ref{inference}), as shown in Fig. \ref{overview}.

\subsection{Text Query Representation} 
\label{text_query_representation}

Given a text query with $N$ words, denoted as $\mathcal{T} = \{t_i\}_{i=1}^{N}$, we first use the pre-trained RoBERTa~\cite{roberta} to extract word features. Then we use a fully-connected (FC) layer to map the word features into a lower-dimensional space. After that, we add learnable positional embeddings to the mapped word features and pass them through a Transformer encoder layer, obtaining a sequence of $d$-dimensional contextualized word embedding vectors $Q \in \mathbb{R}^{N\times d}$. Finally, we use a simple attention pooling on $Q$ to obtain sentence embeddings $q \in \mathbb{R}^d$, similar to \cite{MSSL}.

\subsection{Video Representation with Temporal Consolidation} 
\label{video_representation}

Given an untrimmed video $\mathcal{V} = \{I_i\}_{i=1}^{M_f}$ with $M_f$ frames, we employ a pre-trained 2D or 3D CNN to extract frame features, denoted as $F \in \mathbb{R}^{M_f \times d_o}$, where $d_o$ is frame feature dimension. Following prior arts~\cite{MSSL, GMMFORMER}, we construct multi-granularity video representations, jointly using a frame-level branch and a clip-level branch.

In the frame-level branch, we directly take frame features $F$ as input and use a FC layer to reduce dimension to $d$. Then we equip the GMMFormer block in \cite{GMMFORMER} with our designed temporal consolidation module (TC-GMMBlock) to obtain frame embeddings $V_f = \{f_i\}_{i=1}^{M_f} \in \mathbb{R}^{M_f\times d}$. Compared with the video-level branch in GMMFormer~\cite{GMMFORMER}, our frame-level branch can extract fine-grained local information and reflect differences in video duration.

The clip-level branch first downsamples the input in the temporal domain to integrate frames into clips. Similar to \cite{MSSL}, we uniformly sample a fixed number $M_c$ (\ie 25\% of the total frame amount $M_f$) of clip features by mean pooling over the corresponding multiple consecutive frame features $F$. Then we utilize a FC layer and TC-GMMBlock on clip features to obtain clip embeddings $V_c = \{c_i\}_{i=1}^{Mc} \in \mathbb{R}^{M_c\times d}$, which contain adaptive clip information and help the model to perceive relevant moments.

\subsubsection{Preliminaries: GMMFormer Block}

\setlength{\intextsep}{0pt}
\begin{wrapfigure}{r}{0.45\textwidth}
  \begin{center}
   \includegraphics[width=\linewidth]{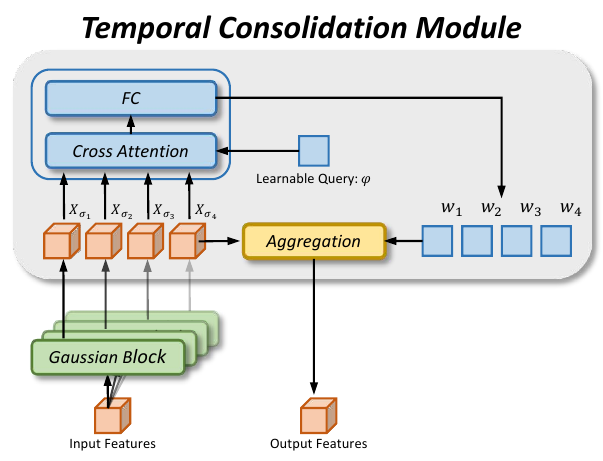}
  \end{center}
  \caption{\textbf{The detailed architecture of TC-GMMBlock.}}
  \label{tcblock}
\end{wrapfigure}
Recently, \cite{GMMFORMER} introduced a Transformer model called GMMFormer for efficient PRVR. Each block in GMMFormer (GMMBlock) incorporates Gaussian-Mixture-Model priors in the contextual encoding process to encourage local modeling capacity. Specifically, GMMBlock is composed of a series of Gaussian-constrained Transformer blocks (Gaussian blocks) and a static aggregation module. Each Gaussian block pre-defines a specific Gaussian matrix to extract features of a specific scale. After extracting multi-scale contextual features in parallel, GMMFormer utilizes average pooling to aggregate them. We provide detailed formulas for GMMBlock in the appendix.

Such a static aggregating approach could potentially introduce irrelevant clip information and reduce the proportion of correct clip information. Furthermore, it might miss the target moment when handling videos that contain unanticipated M/V moments.

\subsubsection{Temporal Consolidation Module}
\label{sec33}

In this subsection, we present TC-GMMBlock to better perceive video moments with varying lengths, depicted in Fig. \ref{tcblock}. Specifically, we improve static aggregation in GMMBlock by designing a novel temporal consolidation module (TCM). Time points in a video may be located at moments of different lengths. Motivated by this, TCM learns adaptive aggregation weights and assigns varying receptive fields for different time points in a video.

Specifically, we define a learnable query $\varphi \in \mathbb{R}^d$ and build the aggregation weight generator with a block-aware cross-attention layer ($\operatorname{CA}$) and a FC layer ($\operatorname{FC}$), which generates aggregation weights for different time points to aggregate multi-scale contextual features:
\begin{gather}
    w_k = \operatorname{FC}(\operatorname{CA}(\varphi, X_{\sigma_{k}}, X_{\sigma_{k}})), k = 1, 2, ..., K, \\
    \Tilde{X}_j =  \sum_{k=1}^{K} \Tilde{w}_{k,j} X_{\sigma_{k},j},\  \Tilde{w}_{k,j} = \frac{e^{w_{k,j} / \tau}}{\sum_{i=1}^{K}e^{w_{i, j} / \tau}}, j = 1, 2, ..., M, \\
    X_{TCM} = \operatorname{concat}(\Tilde{X}_1, \Tilde{X}_2, ..., \Tilde{X}_M),
\end{gather}
where $X_{\sigma_{k}} \in \mathbb{R}^{M\times d}$ is the output of the $k$-th Gaussian block, $K$ is the number of Gaussian blocks and $M$ is the time point number. $w_k \in\mathbb{R}^{M}$ denotes the aggregation weights for the $k$-th Gaussian block and $\tau$ is the temperature factor. $\Tilde{X}_j \in\mathbb{R}^{d}$ denotes the aggregated features at time point $j$ and $X_{TCM}$ is the output of the temporal consolidation module. With the designed temporal consolidation module, the model can better perceive varying uncertain moments and improve the video representations.

\subsection{Uncertainty-aware Text-clip Matching}
\label{uncertainty_aware_cross_modal_matching}

\begin{figure}[t]
  \centering
  \includegraphics[width=\linewidth]{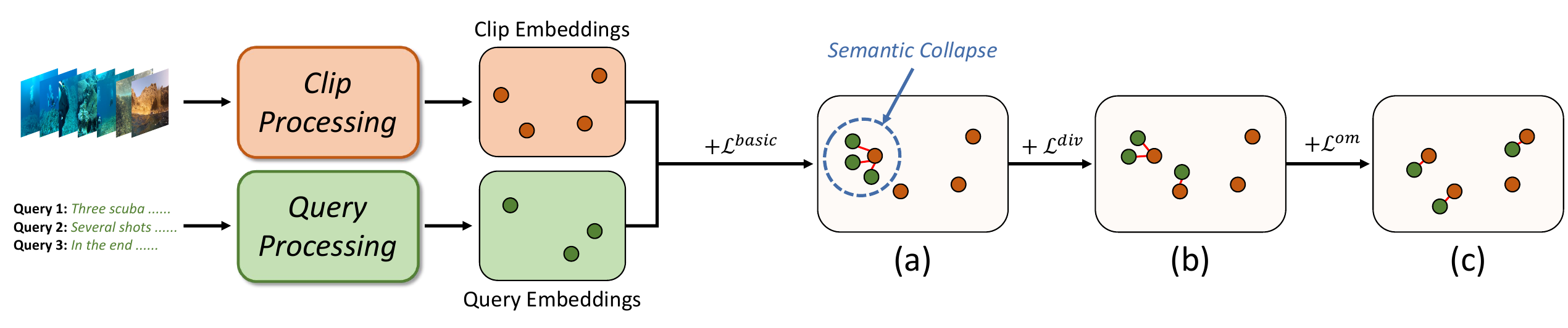}
  \caption{\textbf{The semantic collapse problem and our solution.} (a) With only basic retrieval training loss $\mathcal{L}^{basic}$, we find a semantic collapse phenomenon.
  (b) Query diverse loss $\mathcal{L}^{div}$ can preserve the text semantic structure by encouraging fine-grained \emph{uniformity}.
  (c) Optimal matching loss $\mathcal{L}^{om}$ can assure non-redundant matching between text queries and relevant clips, neatly promoting fine-grained text-clip \emph{alignment}. 
  Red edges between text queries and clips reflect the optimal assignments.}
  \label{loss_ill}
\end{figure}

\subsubsection{Phenomenon: Semantic Collapse} During training, we find a semantic collapse phenomenon, where the semantic structure of text representations is disturbed and different text queries tend to be assigned to a few overlapped clips within the video as shown in Fig. \ref{loss_ill}(a). This will result in most truly relevant clip embeddings not receiving supervision signals and degrade the cross-modal learning. To solve this problem, we revamp the query diverse loss in GMMFormer to preserve a better text semantic structure and propose an optimal matching loss to enhance the diversity of clip embeddings for text queries.

\subsubsection{Revamped Query Diverse Loss}
\label{query_diverse_loss}

An untrimmed video typically contains various moments with diverse semantics, which can be reflected by different text queries. Therefore, \cite{GMMFORMER} argued that different text queries should preserve fine-grained discriminative structure and proposed a query diverse loss to push away texts relevant to the same video in the embedding space during training. However, the pushing degree of different text pairs is identical. For those well-preserved text pairs\footnote{A text pair is well-preserved when the embedding distance of two texts in the semantic space already matches with their semantic relationship.}, keeping pushing them will conflict with retrieval training. 
To overcome this problem, we revamp the query diverse loss by applying different pushing strengths to different text pairs. In particular, we propose to reshape the loss function to down-weight well-preserved text pairs and thus focus training on hard text pairs. More formally, 
we equip a modulating factor $(1 + \operatorname{cos}(q_i, q_j))^{\gamma}$ to the query diverse loss \wrt the relevant text pair $(q_i, q_j)$, where $\gamma > 0$ is a tunable focusing factor. Accordingly, we define the revamped query diverse loss as:
\begin{gather}
\ell(i,j) = (1 + \operatorname{cos}(q_i, q_j))^{\gamma} \operatorname{log}(1 + e^{\alpha (\operatorname{cos}(q_i, q_j) + \delta)}), \label{eq_gamma} \\
\mathcal{L}^{div} = \frac{2}{M_q (M_q - 1)} \sum_{1 \leq i, j \leq M_q, i \neq j} \ell(i,j),
\end{gather}
where $\delta > 0$ is a margin factor,
$\alpha > 0$ is a scaling factor and 
$M_q$ is the number of text queries relevant to a video. 
During the learning process, the revamped query diverse loss plays a promoting role for the fine-grained uniformity of the semantic structure, as illustrated in Fig. \ref{loss_ill}(b).

\subsubsection{Optimal Matching Loss}
\label{optimal_matching_loss}

In practice, we observe that text queries relevant to the same video tend to be assigned to a few homogeneous clips, which will result in most truly relevant clip embeddings not receiving supervision signals and degrade the cross-modal learning. To address this problem, we consider the assignment as a maximum matching problem on an unweighted bipartite graph, where clip embeddings $\{c_i\} \in \mathbb{R}^{M_c\times d}$ of a video and its relevant text query embeddings $\{q_i\} \in \mathbb{R}^{M_q\times d}$ form two vertex sets. 
We encode all the text-clip similarity scores \wrt this video to represent the profits of the edges on the bipartite graph, namely $\Pi = [\pi_{ij}]_{M_q \times M_c}$, $\pi_{ij} = \operatorname{cos}(q_{i}, c_j)$, and we assume that $M_q \le M_c$ in practice. 
Then, our goal is to find the optimal non-overlapping assignment plan $A^*=[a_{ij}^*]_{M_q\times M_c}$, $a_{ij} \in \{0, 1\}$, which maximizes the overall matching profits: 
\begin{gather}
    A^* = \underset{A}{{\arg\max}} \sum_{i,j} \pi_{ij} a_{ij},\quad  
    \textrm{s.t.}\ 
    \forall i: \sum_{j} a_{ij} = 1.
\end{gather}

It can be easily solved in polynomial time using the Hungarian algorithm~\cite{kuhn1955hungarian}.
We align text-clip pairs \wrt each video in the embedding space according to the optimal assignment, yielding the optimal matching loss:
\begin{gather}
    \mathcal{L}^{om} = 
    \frac{1}{M_q} \sum_{i,j} (1-\operatorname{cos}(q_{i}, c_j)) \cdot a_{ij}^{*}.
\end{gather}

The optimal matching loss ensures the diversity of text-clip matching and neatly facilitates fine-grained alignment under uncertainty, as shown by Fig. \ref{loss_ill}(c).

\subsubsection{Total Training Objectives}
\label{sec36}
Firstly, we adopt basic loss $\mathcal{L}^{basic}$ used in GMMFormer~\cite{GMMFORMER} for retrieval training. Then, we utilize the above-mentioned query diverse loss $\mathcal{L}^{div}$ and optimal matching loss $\mathcal{L}^{om}$ for uncertainty-aware text-clip matching. To sum up, we train the model to minimize:
\begin{gather}
\mathcal{L} = \mathcal{L}^{basic} + \lambda_d \mathcal{L}^{div} + \lambda_o \mathcal{L}^{om},
\end{gather}
where $\lambda_d$ and $\lambda_o$ are adopted to balance the different regularization terms.

\subsection{Similarity Measure} 
\label{inference}

To measure the similarity between a text-video pair $(\mathcal{T}, \mathcal{V})$, we first calculate $q$, $V_f$ and $V_c$ according to Sec. \ref{text_query_representation} and Sec. \ref{video_representation}. 
Then, we compute the frame-level and clip-level similarity scores respectively: 
\begin{gather}
S_f(\mathcal{T}, \mathcal{V}) = \operatorname{max} \{\operatorname{cos}(q, f_1), \operatorname{cos}(q, f_2), ..., \operatorname{cos}(q, f_{M_f})\}, \\
S_c(\mathcal{T}, \mathcal{V}) = \operatorname{max} \{\operatorname{cos}(q, c_1), \operatorname{cos}(q, c_2), ..., \operatorname{cos}(q, c_{M_c})\}.
\end{gather}

Next, we interpolate the two scores to obtain the overall similarity: 
\begin{gather}
S(\mathcal{T}, \mathcal{V}) = \alpha_f S_f(\mathcal{T}, \mathcal{V}) + \alpha_c S_c(\mathcal{T}, \mathcal{V}),
\end{gather}
where $\alpha_f, \alpha_c \in [0, 1]$ are pre-defined interpolation factors satisfying $\alpha_f + \alpha_c = 1$.
Finally, given a text query, we retrieve and rank partially relevant videos according to the computed similarities.

\section{Experiments}

\subsection{Experimental Setup}

\subsubsection{Datasets and Metrics}

We evaluate our GMMFormer v2 on three PRVR benchmark datasets, namely ActivityNet Captions~\cite{activitynet}, TV show Retrieval (TVR)~\cite{tvr} and Charades-STA~\cite{sta}. Notably, moment annotations provided by these datasets are unavailable in the PRVR task. Following prior arts~\cite{MSSL}, we use rank-based metrics to evaluate retrieval performance, \ie R$K$ (Recall@$K$) where $K$ is equal to 1, 5, 10 and 100, and Sum of all Recalls (SumR). More details are provided in the appendix.

\begin{table}[t]
\centering
\caption{\textbf{Retrieval results of GMMFormer v2 and other SOTA methods on ActivityNet Captions, TVR and Charades-STA.} Models are sorted in ascending order in terms of their overall performance on ActivityNet Captions. State-of-the-art performance is marked in bold and the underlined numbers are the second best performances. “-” means that the corresponding results are not available.}
\label{sota}
\scalebox{0.64}{
\begin{tabular}{l|ccccccc|cccccc|cccccc}\toprule
\multirow{2}{*}{Method}&&\multicolumn{6}{c|}{ActivityNet Captions} &\multicolumn{6}{c|}{TVR} &\multicolumn{6}{c}{Charades-STA} \\
\cmidrule(lr){2-8} \cmidrule(lr){9-14} \cmidrule(lr){15-20}
 &&R1&R5&R10&R100&&SumR&R1&R5&R10&R100&&SumR&R1&R5&R10&R100&&SumR \\\cmidrule{1-20}
 DE~\cite{dong2019dual} && 5.6 & 18.8 & 29.4 & 67.8 && 121.7 & 7.6 & 20.1 & 28.1 & 67.6 && 123.4      & 1.5 & 5.7 & 9.5 & 36.9 && 53.7 \\
 W2VV++~\cite{li2019w2vv++} && 5.4 & 18.7 & 29.7 & 68.8 && 122.6 & 5.0 & 14.7 & 21.7 & 61.8 && 103.2      & 0.9 & 3.5 & 6.6 & 34.3 && 45.3 \\
 CE~\cite{liu2019use} && 5.5 & 19.1 & 29.9 & 71.1 && 125.6 & 3.7 & 12.8 & 20.1 & 64.5 && 101.1      & 1.3 & 4.5 & 7.3 & 36.0 && 49.1 \\
 ReLoCLNet~\cite{zhang2021video} && 5.7 & 18.9 & 30.0 & 72.0 && 126.6 & 10.7 & 28.1 & 38.1 & 80.3 && 157.1      & 1.2 & 5.4 & 10.0 & 45.6 && 62.3 \\
 XML~\cite{tvr} && 5.3 & 19.4 & 30.6 & 73.1 && 128.4 & 10.0 & 26.5 & 37.3 & 81.3 && 155.1      & 1.6 & 6.0 & 10.1 & 46.9 && 64.6 \\
 MS-SL~\cite{MSSL} && 7.1 & 22.5 & 34.7 & 75.8 && 140.1 & 13.5 & 32.1 & 43.4 & 83.4 && 172.4      & 1.8 & 7.1 & 11.8 & 47.7 && 68.4 \\
 JSG~\cite{JSG} && 6.8 & 22.7 & 34.8 & 76.1 && 140.5 & - & - & - & - && -      & 2.4 & 7.7 & 12.8 & 49.8 && 72.7 \\
  UMT-L~\cite{li2023unmasked} && 6.9 & 22.6 & 35.1 & 76.2 && 140.8 & 13.7 & 32.3 & 43.7 & 83.7 && 173.4      & 1.9 & 7.4 & 12.1 & 48.2 && 69.6 \\
 PEAN~\cite{PEAN} && 7.4 & 23.0 & 35.5 & 75.9 && 141.8 & 13.5 & 32.8 & 44.1 & 83.9 && 174.2      & \textbf{2.7} & \underline{8.1} & \underline{13.5} & 50.3 && \underline{74.7} \\
 InternVideo2~\cite{wang2024internvideo2} && 7.5 & 23.4 & 36.1 & 76.5 && 143.5 & 13.8 & 32.9 & 44.4 & 84.2 && 175.3      & 1.9 & 7.5 & 12.3 & 49.2 && 70.9 \\
 GMMFormer~\cite{GMMFORMER} && \underline{8.3} & 24.9 & 36.7 & 76.1 && 146.0 & 13.9 & 33.3 & 44.5 & \underline{84.9} && 176.6      & 2.1 & 7.8 & 12.5 & \underline{50.6} && 72.9 \\
 DL-DKD~\cite{DKD} && 8.0 & \underline{25.0} & \underline{37.5} & \underline{77.1} && \underline{147.6} & \underline{14.4} & \underline{34.9} & \underline{45.8} & \underline{84.9} && \underline{179.9}      & - & - & - & - && - \\
\rowcolor[gray]{.92}
 \textbf{Ours} && \textbf{8.9} & \textbf{27.1} & \textbf{40.2} & \textbf{78.7} && \textbf{154.9} & \textbf{16.2} & \textbf{37.6} & \textbf{48.8} & \textbf{86.4} && \textbf{189.1}      & \underline{2.5} & \textbf{8.6} & \textbf{13.9} & \textbf{53.2} && \textbf{78.2} \\
\bottomrule
\end{tabular}
}
\end{table}

\setlength{\intextsep}{0pt}
\begin{wraptable}{r}{0.45\textwidth}
\centering
\caption{\textbf{Model comparisons in terms of FLOPs, parameters and retrieval efficiency.} For a fair comparison, the reported runtime is measured on the same Nvidia RTX3080Ti GPU.}
\label{com}
\scalebox{0.62}{
\begin{tabular}{l|cccccc}\toprule
  && MS-SL && GMMFormer && GMMFormer v2 \\
\cmidrule{1-7}
FLOPs (G) && 1.29 && 1.95 && 5.43 \\
Params (M) && 4.85 && 12.85 && 32.27 \\
Runtime (ms) && 12.93 && 4.56 && 6.73 \\
Memory (M) && 250.11 && 12.67 && 61.57 \\
\bottomrule
\end{tabular}}
\end{wraptable}
\subsubsection{Implementation Details}

We use I3D features from \cite{zhang2020hierarchical} on ActivityNet Captions and Charades-STA for video representations. On TVR, we concatenate frame-level ResNet152~\cite{he2016deep} features and segment-level I3D~\cite{carreira2017quo} features to obtain 3,072-D visual features, as provided by \cite{tvr}. For sentence representations, we use 1,024-D RoBERTa features provided by \cite{MSSL} on ActivityNet Captions and Charades-STA, 768-D RoBERTa features provided by \cite{tvr} on TVR. In TC-GMMBlock, we employ eight Gaussian blocks and set the Gaussian variances to 0.1, 0.5, 1.0, 3.0, 5.0, 8.0, 10.0 and $\infty$, respectively. Due to space constraints, we provide more implementation details in the appendix.

\subsection{Comparison with State-of-the-art Methods}

We conduct comparisons of our GMMFormer v2 against competitive models designed for text-to-video retrieval (T2VR)~\cite{dong2019dual, li2019w2vv++, liu2019use}, video corpus moment retrieval (VCMR)~\cite{zhang2021video, tvr, JSG}, general video understanding (GVU)~\cite{li2023unmasked, wang2024internvideo2} and partially relevant video retrieval (PRVR)~\cite{MSSL, PEAN, DKD, GMMFORMER}. 

As shown in Tab. \ref{sota}, T2VR models perform poorly due to the lack of ability to perceive untrimmed videos. With moment supervision, VCMR models go one step further than T2VR models. Besides, GVU models pretrained on web-scale text-video pairs perform well. However, their performance is still inferior to PRVR models due to the lack of clip modeling.

Among PRVR models, our GMMFormer v2 surpasses previous works by a significant margin. Specifically, GMMFormer v2 achieves \textbf{6.1\%}, \textbf{7.1\%} and \textbf{7.3\%} relative lift in SumR than GMMFormer on three benchmarks, respectively, and surpasses the past SOTA competitor, DL-DKD, which utilizes the rich generalization knowledge from pretrained vision-language models. Considering the particularity and difficulty of the PRVR task, such performance improvements are drastic.

Compared with previous SOTA methods, the major advantage of GMMFormer v2 lies in better capacity in capturing video moments with varying lengths. With the designed temporal consolidation module, the model can better locate correct video moments relevant to given text queries.

\subsection{Complexity Analyses}

In this section, we provide complexity analyses of our GMMFormer v2. To reflect the retrieval efficiency in an actual situation, we first build a database of 2,500 videos from TVR. Then we choose several competitive models to measure their retrieval efficiency, including FLOPs, parameters, runtime and memory usage to complete the retrieval process for a single text query. As shown in Tab. \ref{com}, the introduction of the temporal consolidation module and more extensive Gaussian blocks indeed increases the computational complexity of GMMFormer v2. However, those additional calculations are located in video branches, which will be offline to compute beforehand. In the actual inference, GMMFormer v2 is efficient (\ie about 2 times faster than the classic PRVR model MS-SL, and the storage overhead is 4 times smaller than MS-SL).

\subsection{Ablations and Analyses}

We present ablation studies on the designed frame-level branch, TC-GMMBlock, revamped query diverse loss, proposed optimal matching loss. We mainly report the results on ActivityNet Captions.

\subsubsection{Fusion Components}

As shown in Tab. \ref{ablation_1}, starting with a pure baseline (Exp.1), GMMFormer v2 gains \textbf{2.7\%} on SumR by replacing the video-level branch with the frame-level branch (Exp.2). Then, GMMFormer v2 gains \textbf{3.8\%} on SumR by replacing the vanilla Transformer encoder layer with the designed TC-GMMBlock (Exp.3). Adding revamped query diverse loss further brings an improvement to \textbf{9.3\%} (Exp.4) compared to baseline. By jointly using our designed frame-level branch, TC-GMMBlock, query diverse loss and optimal matching loss, GMMFormer v2 acquires an improvement of \textbf{10.9\%} on SumR (Exp.5). These ablations demonstrate the effectiveness of our designed components in improving retrieval performance.

\begin{table}[t]
\centering
\caption{\textbf{Ablation studies of the fusion components on ActivityNet Captions.} FB, TC-GB, QDL and OM stand for the frame-level branch, TC-GMMBlock, revamped query diverse loss and designed optimal matching loss, respectively.}
\label{ablation_1}
\scalebox{0.8}{
\begin{tabular}{c|cccc|ccccccc}\toprule
Exp. & FB & TC-GB & QDL & OM &&R1&R5&R10&R100&&SumR \\
\cmidrule{1-12}
1   & & & & && 6.8 & 22.8 & 34.8 & 75.3 && 139.7 \\
2   & \checkmark & & & && 7.5 & 23.5 & 35.8 & 76.7 && 143.5 \\
 3 & \checkmark & \checkmark & & && 8.2 & 25.5 & 38.1 & 77.1 && 149.0 \\
  4 & \checkmark & \checkmark & \checkmark & && 8.6 & 26.3 & 39.5 & 78.4 && 152.7 \\
\cmidrule{1-12}
\rowcolor[gray]{.92}
  5 & \checkmark & \checkmark & \checkmark & \checkmark && \textbf{8.9} & \textbf{27.1} & \textbf{40.2} & \textbf{78.7} && \textbf{154.9} \\
\bottomrule
\end{tabular}}
\end{table}

\setlength{\intextsep}{0pt}
\begin{wraptable}{r}{0.45\textwidth}
\centering
\caption{\textbf{Comparison of different aggregation approaches on ActivityNet Captions.} AP, WA, DA and TCM correspond to the average pooling, weighted aggregation, dynamic aggregation and temporal consolidation module.}
\label{ablation_2}
\scalebox{0.7}{
\begin{tabular}{l|ccccccc}\toprule
Aggregation  &&R1&R5&R10&R100&&SumR \\
\cmidrule{1-8}
AP && 8.5 & 25.5 & 37.9 & 77.1 && 149.0 \\
 WA && 8.7 & 26.2 & 38.8 & 78.0 && 151.7 \\
 DA  && 8.6 & 26.6 & 39.3 & 78.3 && 152.8\\
 \cmidrule{1-8}
\rowcolor[gray]{.92}
  TCM  && \textbf{8.9} & \textbf{27.1} & \textbf{40.2} & \textbf{78.7} && \textbf{154.9} \\
\bottomrule
\end{tabular}}
\end{wraptable}
\subsubsection{Aggregation Strategy}

Various types of aggregation strategies can be used in our GMMFormer v2. We explore several representative methods to choose the best one for our model: average pooling, weighted aggregation, dynamic aggregation and our designed temporal consolidation module. As shown in Tab. \ref{ablation_2}, average pooling performs the worst. The reason is that when the number of Gaussian blocks is large (\eg 8), average pooling will introduce a wide range of irrelevant clip information, reducing the proportion of relevant clip information. Weighted aggregation is better than average pooling, demonstrating its effectiveness by highlighting essential Gaussian blocks. Dynamic aggregation goes one step further by generating different aggregation weights for different videos. And our designed temporal consolidation module achieves the best performance, due to its ability in learning adaptive aggregation weights and assigning varying receptive fields for different time points in a video. More details are provided in the appendix.

\subsubsection{Impact of the Focusing Parameter $\gamma$}

 The focusing parameter $\gamma$ smoothly adjusts the rate at which well-preserved text pairs are down-weighted. Fig. \ref{trip}(a) shows the impact of $\gamma$ in Eq. \ref{eq_gamma}. It is observed that as $\gamma$ increases, the performance increases at first and then drops. This is because when $\gamma$ is small, the pushing degree of query diverse loss tends to be identical, leading to possible conflict with retrieval training. In contrast, when $\gamma$ is large, it enhances the pushing degree of query diverse loss to most text pairs, which will drown out retrieval training.

\subsubsection{Impact of the Gaussian Variance $\sigma$}

\begin{figure} [t]
  \centering
  \subfloat[]{\includegraphics[width = 0.5\textwidth]{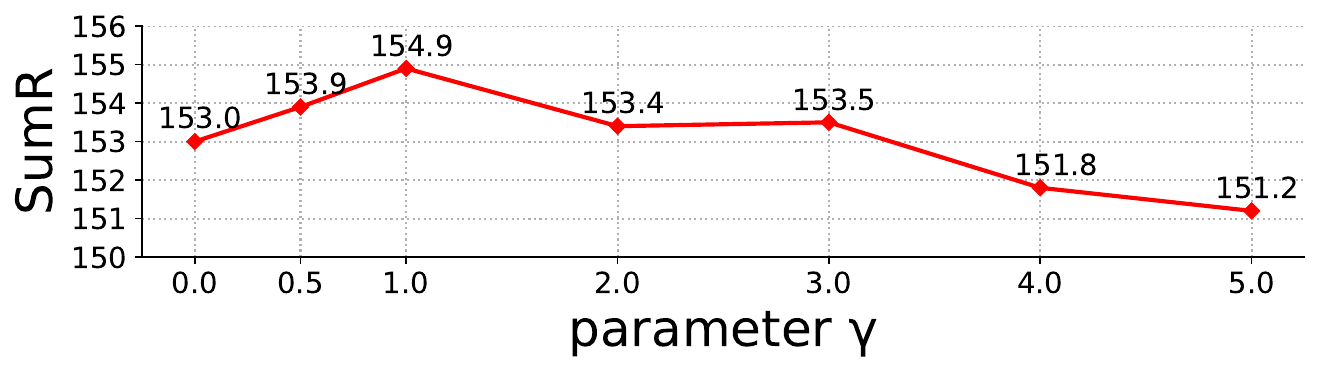}}
  \subfloat[]{\includegraphics[width = 0.5\textwidth]{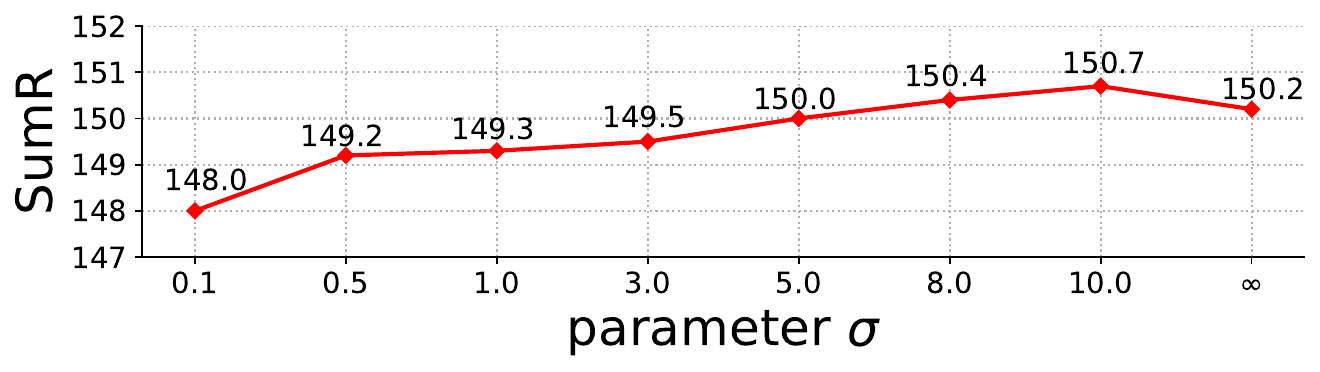}}
  \caption{\textbf{(a) The impact of the hyper-parameter $\gamma$ in Eq. \ref{eq_gamma} and (b) the impact of the Gaussian variance $\sigma$ on ActivityNet Captions.} $\sigma = \infty$ means the vanilla Transformer encoder layer.}
  \label{trip}
\end{figure}

Gaussian variance $\sigma$ controls the local receptive field of input features. In this subsection, we investigate the impact of the Gaussian variance $\sigma$ with a single Gaussian block in GMMFormer v2. As shown in Fig. \ref{trip}(b), larger $\sigma$ generally performs better. The reason may be that when $\sigma$ is small, the model has a small receptive field during feature interactions and fails to explore the temporal dynamics in videos. On the contrary, a large $\sigma$ can result in a large receptive field, which may introduce irrelevant clip information. However, the model can still learn to filter out this useless information to some extent. Notably, TC-GMMBlock is able to capture multi-scale clip information, surpassing all single Gaussian block variants.

\setlength{\intextsep}{0pt}
\begin{wraptable}{r}{0.45\textwidth}
\centering
\caption{\textbf{The impact of the number of Guassian blocks on ActivityNet Captions.} }
\label{number}
\scalebox{0.62}{
\begin{tabular}{l|cccccccc}\toprule
  && 2 && 4 && 6 && 8 \\
\cmidrule{1-9}
GMMFormer && 142.2 && 146.0 && 145.4 && 144.6 \\
GMMFormer v2 && 151.0 && 152.5 && 153.8 && 154.9 \\
\bottomrule
\end{tabular}}
\end{wraptable}
\subsubsection{Impact of the Number of Gaussian Blocks}
\label{gb_number}
In this subsection, we investigate the impact of the number of Gaussian blocks in GMMFormer v2. We uniformly sample different numbers of Gaussian blocks and provide their SumR results in Tab. \ref{number}. For GMMFormer, as the Gaussian block number increases, the performance initially increases and then drops. For GMMFormer v2, the performance is always better than GMMFormer and continues to increase, showing its scalability in utilizing various Gaussian constraints due to the designed temporal consolidation module.

\setlength{\intextsep}{0pt}
\begin{wraptable}{r}{0.45\textwidth}
\centering
\caption{\textbf{The results of the versatility experiment on two classical PRVR models.} }
\label{general}
\scalebox{0.62}{
\begin{tabular}{l|cccccc}\toprule
  && original && w/ $\mathcal{L}^{div}$ && w/ $\mathcal{L}^{div}$ and $\mathcal{L}^{om}$ \\
\cmidrule{1-7}
MS-SL && 140.1 && 144.3 ( $\color{red}{+4.2}$ ) && 146.6 ( $\color{red}{+6.5}$ ) \\
DL-DKD && 147.6 && 150.8 ( $\color{red}{+3.2}$ ) && 152.2 ( $\color{red}{+4.6}$ ) \\
\bottomrule
\end{tabular}}
\end{wraptable}
\subsubsection{Versatility of Uncertainty-aware Text-clip Matching}
\label{sec_general}
To test the versatility of the proposed uncertainty-aware text-clip matching for PRVR, we conduct \textit{plugin-in} experiments on classical PRVR models (MS-SL~\cite{MSSL} and DL-DKD~\cite{DKD}) by adding our query diverse loss $\mathcal{L}^{div}$ and optimal matching loss $\mathcal{L}^{om}$. The SumR results on ActivityNet Captions listed in Tab. \ref{general} show the effective retrieval improvement with the aid of our query diverse loss and optimal matching loss for PRVR models.

\section{Conclusions}
In this paper, we present GMMFormer v2, an uncertainty-aware framework for PRVR. 
To better capture video moments with varying lengths, we improve GMMFormer~\cite{GMMFORMER} with the designed temporal consolidation module to learn adaptive aggregation weights and assign varying receptive fields for different time points in a video.
To better tackle the uncertain correspondence between text queries and moments, we revamp the query diverse loss in GMMFormer to obtain a uniform and discriminative semantic structure and propose an optimal matching loss for fine-grained alignment between text queries and relevant clips. 
They can be used as a \emph{plugin-in} solution to alleviate the semantic collapse issue and neatly promote cross-modal learning for PRVR. 
Extensive experiments and ablation studies on three PRVR benchmarks demonstrate the effectiveness of the proposed framework.

\bibliographystyle{plainnat}
\bibliography{main}


\appendix
\clearpage
\newpage

\noindent\textbf{\large{Appendix}}

\section{Implementation Details}

\subsection{Architecture}

\subsubsection{Simple Attention Pooling}

When representing text queries, we utilize a simple attention pooling module in the last step to aggregate contextualized word embedding vectors into compact sentence embeddings, which can be formulated as:
\begin{gather}
    q = \operatorname{SAP}(Q) = \sum_{i=1}^{N} l_i \times q_{t_i}, \quad  l = \operatorname{softmax}(b Q^T),
\end{gather}
where $Q = \{q_{t_i}\}_{i=1}^{N} \in \mathbb{R}^{N\times d}$ is the contextualized word embedding vectors. $N$ is the sequence length and $d$ is the dimension. $b \in \mathbb{R}^{1\times d}$ is a trainable vector and  $l \in \mathbb{R}^{1\times N}$ indicates the attention vector. $\operatorname{SAP}$ is the simple attention pooling module and its output is denoted as $q$.

\subsubsection{Detailed Formulas for GMMBlock}

GMMBlock in GMMFormer~\cite{GMMFORMER} is composed of a series of Gaussian-constrained Transformer blocks (Gaussian blocks) and a static aggregation module. Each Gaussian block pre-defines a specific Gaussian matrix to rescale the attention score, which can be formulated as:
\begin{gather}
    \operatorname{GA}(X) = \operatorname{softmax}\left(\mathcal{M}_{\sigma}^{g} \odot \frac{X W^q (X W^k)^T}{\sqrt{d_h}}\right)X W^v, \\
    X^{'}_{\sigma} = \operatorname{GA}(\operatorname{LN}(X)) + X, \\
    X_{\sigma} = \operatorname{FFN}(\operatorname{LN}(X^{'}_{\sigma})) + X^{'}_{\sigma},
\end{gather}
where $\mathcal{M}_{\sigma}^{g} \in \mathbb{R}^{M\times M}$ is the Gaussian matrix and $\mathcal{M}_{\sigma}^{g}(i, j) = \frac{1}{2\pi} e^{-\frac{(j-i)^2}{\sigma^2}}$. $\sigma^2$ is the variance of the Gaussian distribution. $X\in \mathbb{R}^{M\times d}$ is the input feature sequence, where $M$ is the time point number and $d$ is the feature dimension. $W^q, W^k, W^v$ are three parameters to project $X$ to three matrices, query, key and value. $d_h$ is the latent dimension of attention and $\odot$ indicates the element-wise product function. $\operatorname{GA}$ is the Gaussian attention module. $\operatorname{LN}$ means  Layer Normalization~\cite{LN} and $\operatorname{FFN}$ is the Feed-Forward Network, composed of two FC layers. 

After extracting multi-scale contextual features, GMMFormer utilizes average pooling to aggregate them, which can be formulated as:
\begin{gather}
    \Tilde{X} = \frac{1}{K} \sum_{k=1}^{K} X_{\sigma_{k}},
\end{gather}
where $X_{\sigma_{k}}$ is the output of the $k$-th Gaussian block, $K$ is the number of Gaussian blocks and $\Tilde{X}$ is the output of GMMBlock. Such a static aggregating approach could potentially introduce irrelevant clip information and reduce the proportion of correct clip information. Furthermore, it might miss the target moment when handling videos that contain unanticipated M/V moments.

\subsubsection{Weighted Aggregation}

A simple idea to improve aggregation of multi-scale contextual features is to assign independent and learnable aggregation weights to different Gaussian blocks. The model learns to adapt aggregation weights of different Gaussian blocks according to their importance:
\begin{gather}
    \Tilde{X} =  \sum_{k=1}^{K} \Tilde{w}_k X_{\sigma_{k}},\ 
    \Tilde{w}_k = \frac{e^{w_k / \tau}}{\sum_{i=1}^{K}e^{w_i / \tau}},
\end{gather}
where $w_k \in \mathbb{R}^{1}$ denotes the weight for the $k$-th Gaussian block.

\subsubsection{Dynamic Aggregation}
Dynamic Aggregation is another practical idea to aggregate multi-scale contextual features. In detail, we define a learnable query $\varphi \in \mathbb{R}^d$ and build the weight generatior with a Cross Attention layer ($\operatorname{CA}$) and a FC layer ($\operatorname{FC}$) to generate different aggregation weights $w \in \mathbb{R}^{K}$ for different videos: 
\begin{gather}
    \Tilde{X} = \sum_{k=1}^{K} \Tilde{w}_k X_{\sigma_{k}}, \Tilde{w}_k = \frac{e^{w_k / \tau}}{\sum_{i=1}^{K}e^{w_i / \tau}},\ w = \operatorname{FC}(\operatorname{CA}(\varphi, X, X)) .
\end{gather}

\subsection{Training Details}

\subsubsection{Basic Loss}

We follow existing PRVR works~\cite{MSSL,GMMFORMER,DKD} to adopt triplet loss $\mathcal{L}^{trip}$ and InfoNCE loss $\mathcal{L}^{nce}$ for dual branches as the basic objectives:
\begin{gather}
\mathcal{L}^{basic} = \mathcal{L}_{c}^{trip} + \mathcal{L}_{f}^{trip} + \lambda_c \mathcal{L}_{c}^{nce} + \lambda_f \mathcal{L}_{f}^{nce},
\end{gather}
where the subscripts $f$ and $c$ mark the objectives for the frame-level branch and the clip-level branch, respectively. 
All the objectives are computed based on the estimated cross-modal similarities, \ie
$S_c = \operatorname{max}_{m=1}^{M_c} \operatorname{cos}(q, c_m)$ of the clip-level branch and $S_f = \operatorname{max}_{m=1}^{M_f} \operatorname{cos}(q, f_m)$ of the frame-level branch. 
$\lambda_c$ and $\lambda_f$ are hyper-parameters to balance the contributions of InfoNCE objectives.

We define a text-video pair as positive if the video has a moment relevant to the text, and negative if there is no relevant content. Given a positive text-video pair $(\mathcal{T}, \mathcal{V})$, the triplet ranking loss over the mini-batch $\mathcal{B}$ is formulated as:
\begin{gather}
\mathcal{L}^{trip} = \frac{1}{n} \sum_{(\mathcal{T},\mathcal{V}) \in \mathcal{B}} \{\operatorname{max}(0, m + S(\mathcal{T}^-, \mathcal{V}) - S(\mathcal{T}, \mathcal{V})) \nonumber \\ 
    + \operatorname{max}(0, m + S(\mathcal{T}, \mathcal{V}^-) - S(\mathcal{T},\mathcal{V}))\},
\end{gather}
where $m$ is a margin constant. $\mathcal{T}^-$ and $\mathcal{V}^-$ indicate a negative text for $\mathcal{V}$ and a negative video for $\mathcal{T}$, respectively.

Given a positive text-video pair $(\mathcal{T}, \mathcal{V})$, the infoNCE loss over the mini-batch $\mathcal{B}$ is computed as:
\begin{gather}
\mathcal{L}^{nce} = -\frac{1}{n} \sum_{(\mathcal{T},\mathcal{V}) \in \mathcal{B}} \{\operatorname{log}(\frac{S(\mathcal{T},\mathcal{V})}{S(\mathcal{T},\mathcal{V}) + \sum\nolimits_{\mathcal{T}_i^{-} \in \mathcal{N}_\mathcal{T} }S(\mathcal{T}_i^-, \mathcal{V})}) \nonumber\\
+ \operatorname{log}(\frac{S(\mathcal{T},\mathcal{V})}{S(\mathcal{T},\mathcal{V}) + \sum\nolimits_{\mathcal{V}_i^{-} \in \mathcal{N}_\mathcal{V} }S(\mathcal{T}, \mathcal{V}_i^-)}) \},
\end{gather}
where $\mathcal{N}_\mathcal{T}$ denotes all negative texts of the video $\mathcal{V}$ in the mini-batch, and $\mathcal{N}_\mathcal{V}$ denotes all negative videos of the query $\mathcal{T}$ in the mini-batch.

\subsection{Datasets and Metrics}

\textbf{ActivityNet Captions}~\cite{activitynet} comprises about 20,000 videos sourced from YouTube, each with an average of 3.7 moments along with corresponding sentence descriptions. We use the data partitioning method popularized in \cite{zhang2021video, zhang2020hierarchical}. \textbf{TVR}~\cite{tvr} has 21,800 videos collected from six TV shows, with each video associated with five natural language sentences describing different moments in it. We follow the training and testing methodology outlined in \cite{MSSL}; we use 17,435 videos containing 87,175 moments for training and 2,179 videos containing 10,895 moments for testing. \textbf{Charades-STA}~\cite{sta} includes 6,670 videos with 16,128 text queries. Each video contains around 2.4 moments with corresponding text queries on average. We use the official data partition for model training and testing. 

Following prior arts~\cite{MSSL}, we use rank-based metrics R$K$ to measure the proportion of queries that return the desired results in the top $K$ of the ranked list. The performance is reported in percentage (\%). Additionally, we use the Sum of all Recalls (SumR) to provide an overall comparison of retrieval results.

\subsection{More Implementation Details}

\subsubsection{Hyper-parameter}
Notably, we directly inherent most hyper-parameter settings from GMMFormer. In detail, we use $M_c=32$ for downsampling and set the maximum frame number $M_f=128$. If the number of frames exceeds $M_f$, we uniformly downsample it to $M_f$. For sentences, we set the maximum length of query words to $N=64$ for ActivityNet Captions and $N=30$ for TVR and Charades-STA. Any words beyond the maximum length will be discarded. 

Regarding the attention module, we use a hidden size of $d=384$ and four attention heads. During model training, we utilize an Adam optimizer with a mini-batch size of 128 and set the number of epochs to 100. Our model is implemented in Pytorch and trained on an Nvidia RTX3080Ti GPU. For regularization terms of loss functions, we set them so that all loss function values have the same order of magnitude. You can find detailed hyper-parameter settings in Tab. \ref{param}. To adjust the learning rate during training, we use a learning rate adjustment schedule similar to MS-SL~\cite{MSSL}.

\begin{table}[t]
\centering
\caption{\textbf{Hyper-parameter settings of ActivityNet Captions, TVR and Charades-STA.}}
\label{param}
\scalebox{0.82}{
\begin{tabular}{l|cccccc}\toprule
Params  && ActivityNet Captions && TVR && Charades-STA \\
\cmidrule{1-7}
learning rate && 2.5e-4 && 3e-4 && 2e-4 \\
$\alpha_{f}$ && 0.3 && 0.3&& 0.3 \\
$\alpha_{c}$ && 0.7 && 0.7 && 0.7\\
$\alpha$ && 32 && 32 && 32\\
$\delta$ && 0.2 && 0.15 && 0.2\\
$\gamma$ && 1 && 1 && 1 \\
$m$ && 0.2 && 0.1 && 0.2 \\
$\tau$ && 6e-1 && 9e-2 && 6e-1 \\
$\lambda_c$ && 2e-2 && 5e-2 && 2e-2 \\
$\lambda_f$ && 4e-2 && 4e-2 && 4e-2 \\
$\lambda_d$ && 3e-3 && 8e-5 && 3e-3 \\
$\lambda_o$ && 1.1e-1 && 9e-2 && 1e-1 \\
\bottomrule
\end{tabular}}
\end{table}

\section{More Experimental Results}

\subsection{More Ablations on ActivityNet Captions}

\setlength{\intextsep}{0pt}
\begin{wraptable}{r}{0.45\textwidth}
\centering
\caption{\textbf{Comparison of different constraint types on ActivityNet Captions.} CT means the constraint type.}
\label{ablation_3}
\scalebox{0.7}{
\begin{tabular}{l|ccccccc}\toprule
CT  &&R1&R5&R10&R100&&SumR \\
\cmidrule{1-8}
Boxcar && 8.4 & 25.4 & 38.3 & 78.2 && 150.3 \\
 Bartlett && 8.6 & 25.9 & 38.8 & 78.3 && 151.6 \\
\rowcolor[gray]{.92}
  Gaussian  && \textbf{8.9} & \textbf{27.1} & \textbf{40.2} & \textbf{78.7} && \textbf{154.9} \\
\bottomrule
\end{tabular}}
\end{wraptable}

\subsubsection{Choice of Constraint Type}
In the Gaussian block of TC-GMMBlock, there are several constraint types to focus each feature on its adjacent features during interactions. In this subsection, we alternate three types of constraint types (\ie Boxcar, Bartlett, Gaussian) and report their performance on ActivityNet Captions in Tab. \ref{ablation_3}. Among them, the Boxcar type performs the worst, which is consistent with the intuition that video frames should pay more attention to nearer frames. Additionally, the Gaussian type outperforms the Bartlett type. We attribute this to the smooth and natural characteristics of the Gaussian distribution. Gaussian prior helps to model the temporal aspect of videos since it can encapsulate the notion of a central point with a spread, which aligns well with the idea of an event occurring around a central time frame with a certain duration.

\begin{figure}[t]
  \centering
  \includegraphics[width=\linewidth]{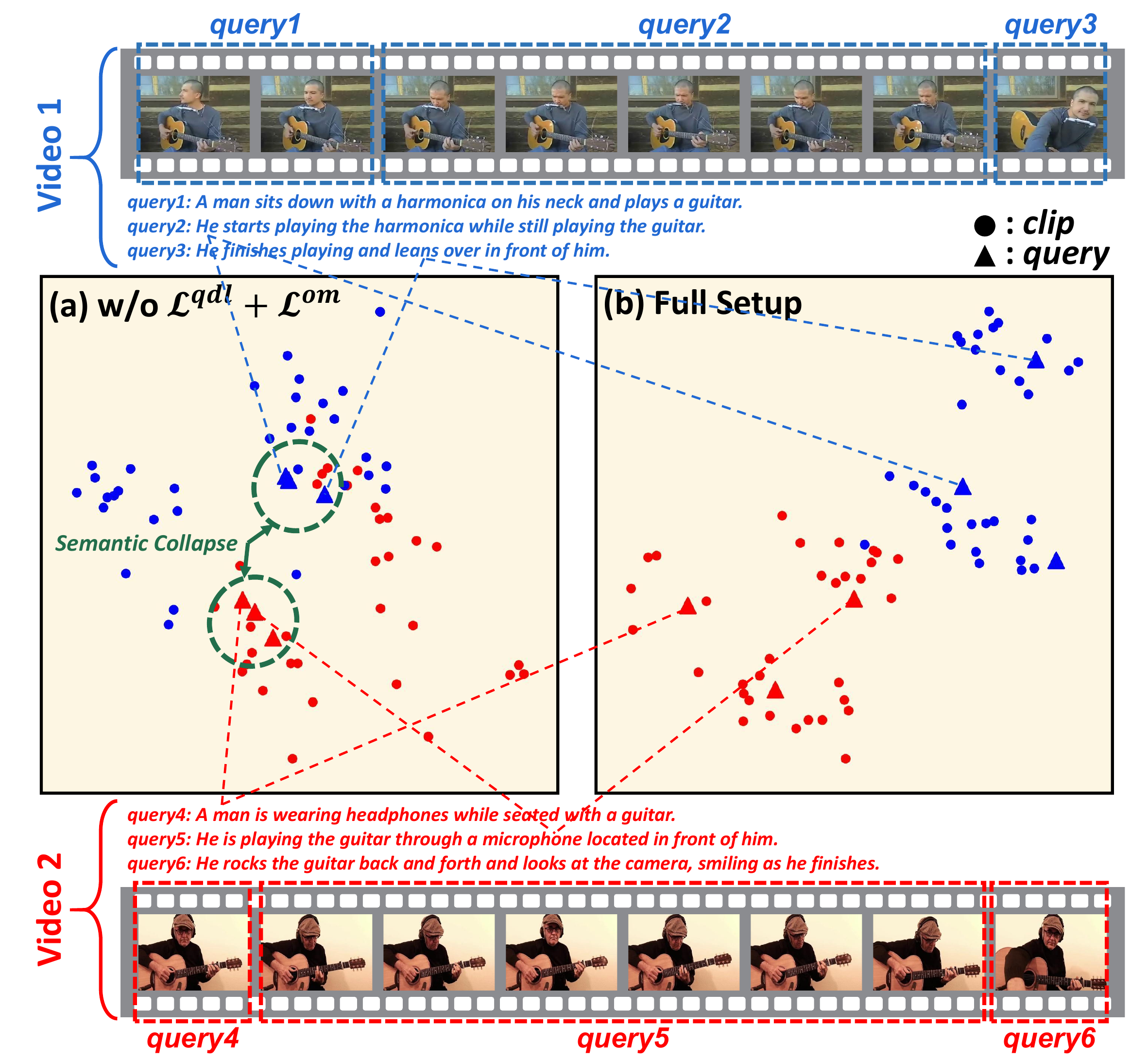}
  \caption{\textbf{t-SNE visualizations of text queries and clip embeddings of two semantically close videos.} $\mathcal{L}^{qdl}$ and $\mathcal{L}^{om}$ represent query diverse loss and optimal matching loss, respectively.} 
  \label{app_vis1}
\end{figure}

\section{Case Study}

To intuitively reveal the causes of semantic collapse and why our proposed combination of query diverse loss and optimal matching loss can alleviate this phenomenon, we randomly select two semantically close videos from ActivityNet Captions by measuring the average embedding distance of relevant text queries and visualize the embedding distribution of their queries and clips by t-SNE, as shown in Fig \ref{app_vis1}. Among them, Fig. \ref{app_vis1}(a) shows the output of a variant of GMMFormer v2 removing query diverse loss and optimal matching loss, Fig. \ref{app_vis1}(b) shows the output of complete GMMFormer v2.

In Fig. \ref{app_vis1}(a), on the one hand, the clip embeddings matched by queries are highly homogeneous (a few points that are close to each other, or even the same point). The effective area in cross-modal matching is only the collapsed small area indicated by the green circle. Due to the lack of moment annotations, the fine-grained text-clip matching pattern is not effectively constrained in this variant, resulting in the model learning shortcuts and causing the semantic collapse problem. On the other hand, most clip embeddings are not activated during training and therefore are not effectively supervised, resulting in clip embeddings belonging to different videos being mixed together and the discriminability of video representations decreasing.

In Fig. \ref{app_vis1}(b), query diverse loss retains more semantic information and makes queries more uniformly distributed in the embedding space. In addition, optimal matching loss makes fine-grained alignment between queries and clips more diverse, helping to reduce noisy cross-modal correspondence caused by collapsed matching. Since diverse clip embeddings are activated during training, through the coarse-grained (video-level) supervision signal of basic loss, clip embeddings belonging to different videos can be made more discriminative, which helps to improve the robustness of PRVR.

\section{Visualizations}

\subsection{Text-clip Similarity Heat Map}
\label{sec_vis}

\begin{figure}[t]
  \centering
  \includegraphics[width=\linewidth]{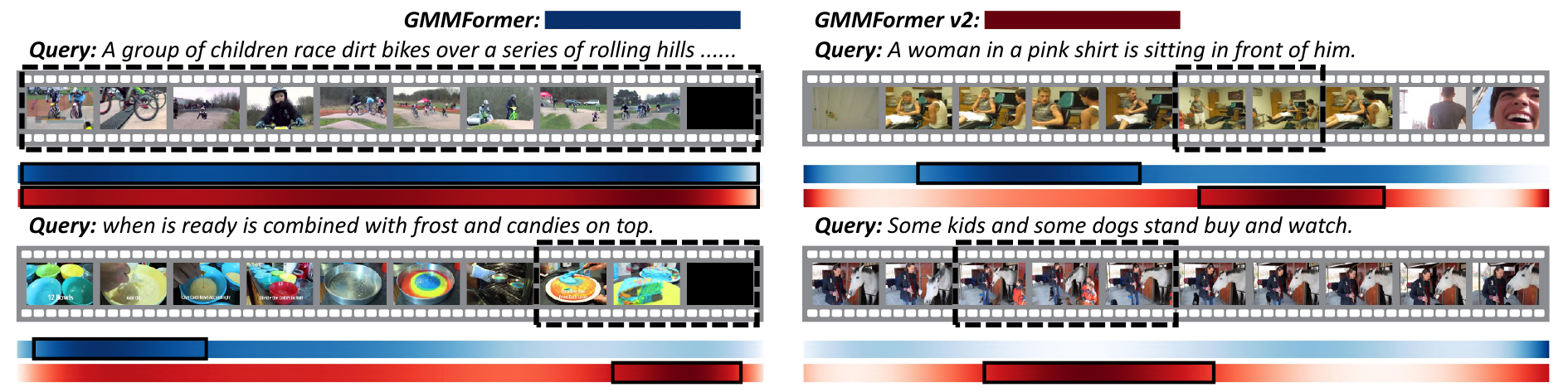}
  \caption{\textbf{Text-clip similarity heat map visualizations of GMMFormer and GMMFormer v2.} The black dotted box represents the target moment area. The darker color area on the heat map is the location the model thinks is more likely to be the target moment and we use the black solid box to represent the area that the model considers most likely to be the target moment. Note that we smooth out text-clip similarity intervals for better observation.}
  \label{vis}
\end{figure}

To further reveal the ability of GMMFormer v2 to perceive target moments in untrimmed videos, we present several text-clip similarity examples on ActivityNet Captions. As shown in Fig. \ref{vis}, compared with GMMFormer, GMMFormer v2 captures the target moment more accurately. For videos with similar scenes or moments with unanticipated M/Vs, GMMFormer v2 exhibits robustness in locating the correct moment while GMMFormer is prone to mistake moments or simply unable to perceive moments. 

\begin{figure} [t]
  \centering
  \subfloat[Random Initialization]{\includegraphics[width = 0.25\textwidth]{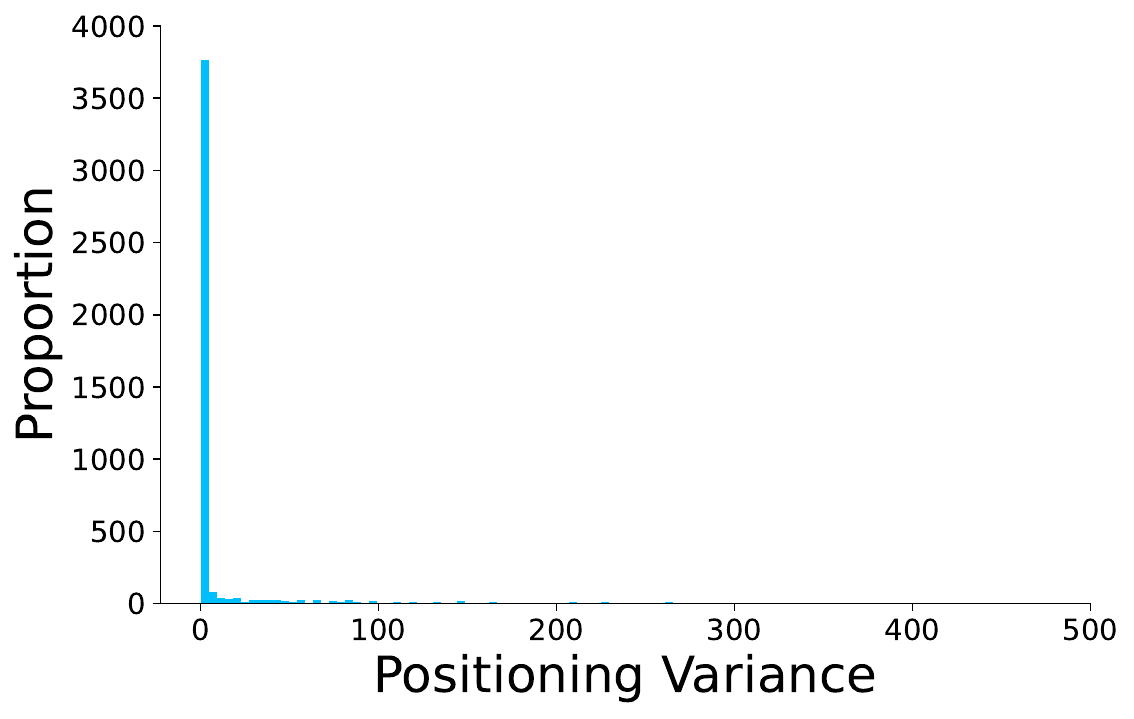}}
  \subfloat[w/o $\mathcal{L}^{qdl}$ and $\mathcal{L}^{om}$]{\includegraphics[width = 0.25\textwidth]{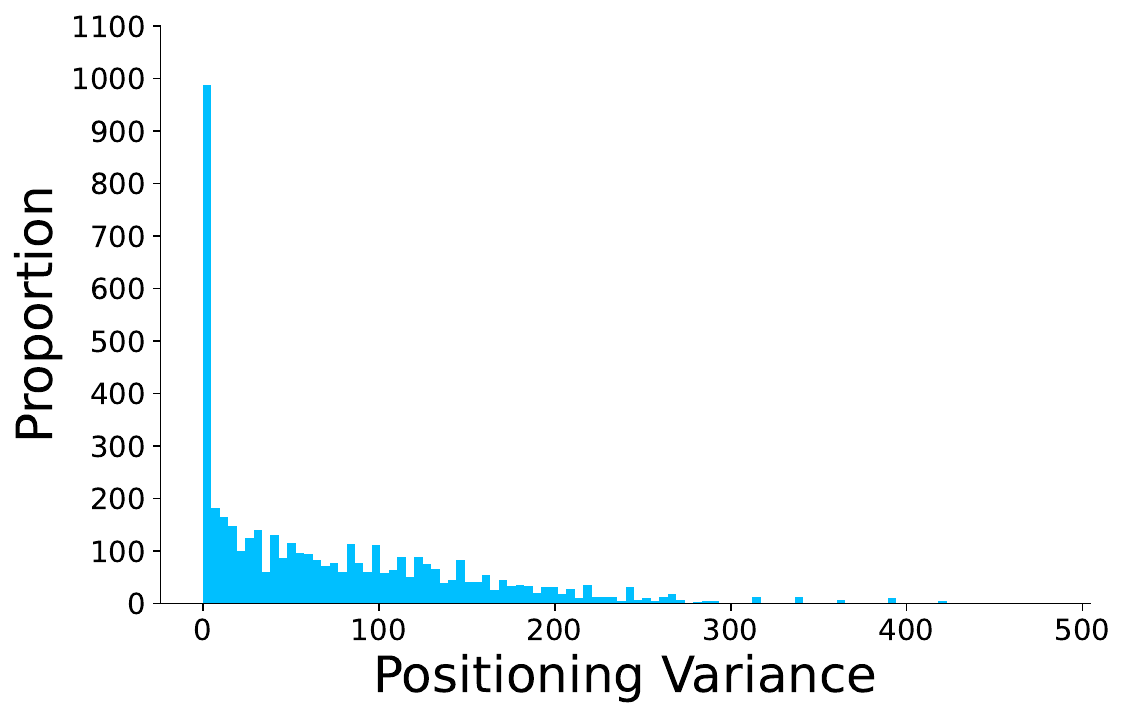}}
  \subfloat[w/o $\mathcal{L}^{om}$]{\includegraphics[width = 0.25\textwidth]{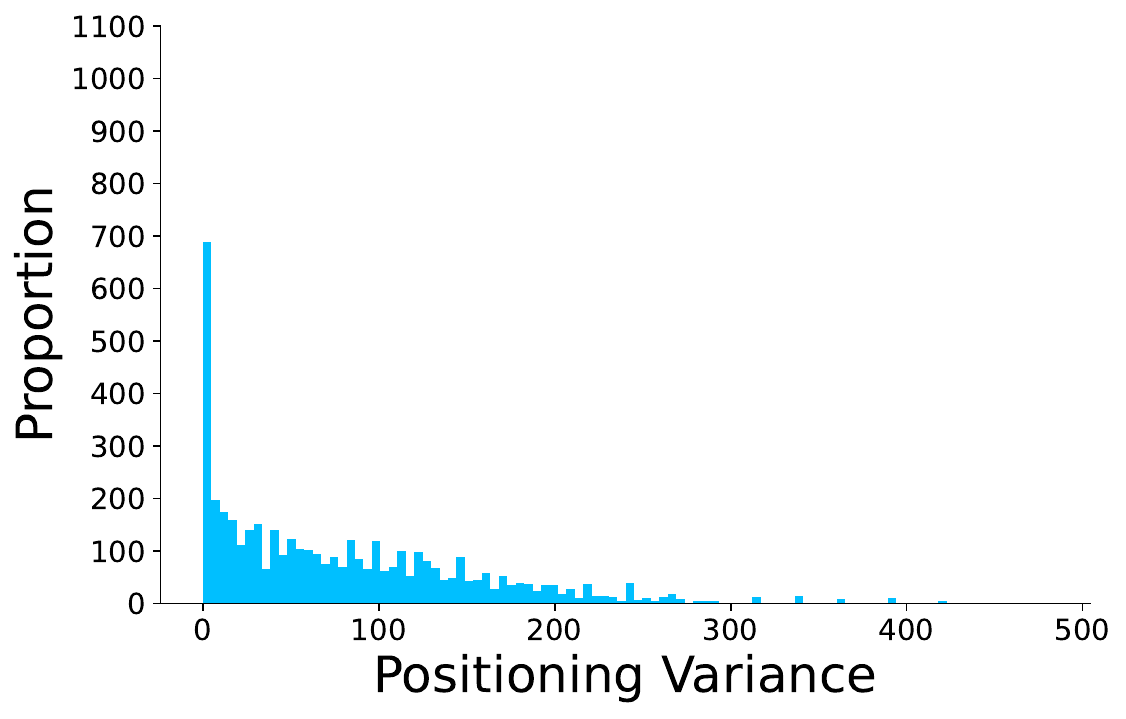}}
  \subfloat[Full Setup]{\includegraphics[width = 0.25\textwidth]{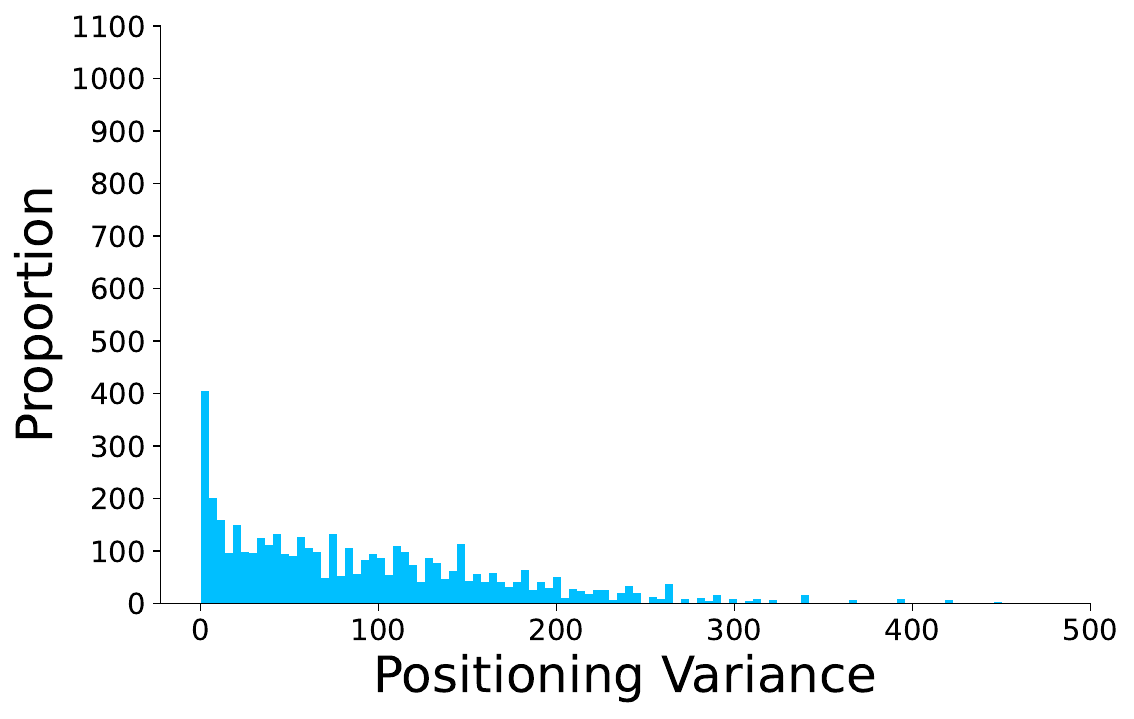}}
  \caption{\textbf{Visualizations of positioning variance distribution on ActivityNet Captions.} Given an untrimmed video and its relevant text queries, we first pass them through our model to obtain clip embeddings and text query embeddings. Then, we obtain the positions of the text query embeddings on the clip embeddings through maximum similarity matching. After that, the positioning variance of this video is defined as the variance of the text query matching positions on clip embeddings.}
  \label{col_vis}
\end{figure}

\subsection{Positioning Variance Distribution} 
\label{vis_2}

In this subsection, we present the distribution of relevant text queries' positioning variance on ActivityNet Captions, which is calculated by the variance between matching positions of relevant text queries on clip embeddings. As shown in Fig. \ref{col_vis}(a), when the model is randomly initialized at the beginning of training, the probability that the positioning variance is close to zero is very high, indicating that different text queries tend to be assigned to a few overlapped clips. By comparing Fig. \ref{col_vis}(b), \ref{col_vis}(c) and \ref{col_vis}(d), we can see that the designed query diverse loss and optimal matching loss both reduce the probability of variance near zero, demonstrating that they can effectively guide different text queries to align to relevant clips within the same video in a fine-grained and diversified manner.

\section{Qualitative Retrieval Results}

\begin{figure}[t]
  \centering
  \includegraphics[width=\linewidth]{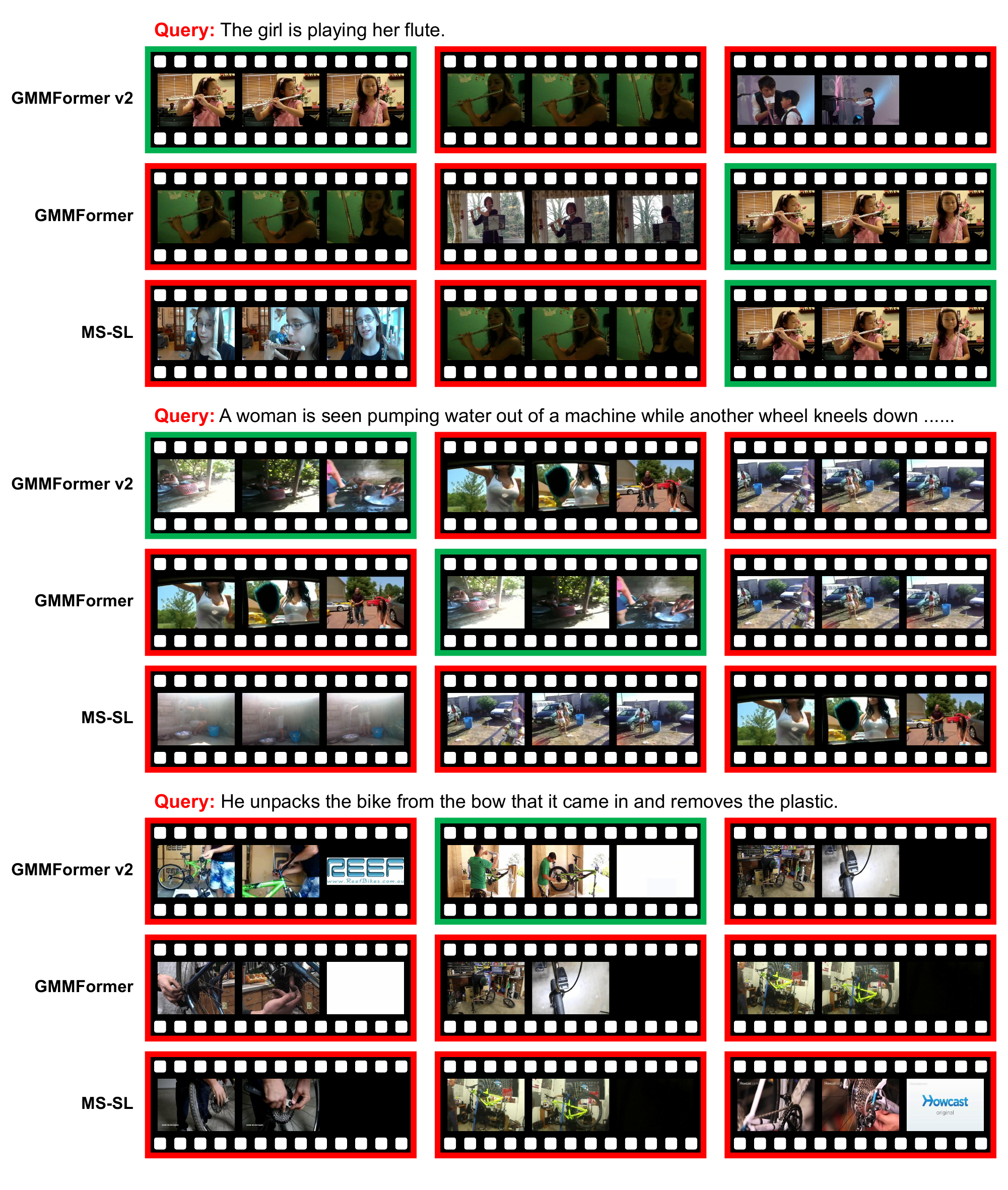}
  \caption{\textbf{Text-to-video retrieval results on ActivityNet Captions.} Top three retrieval results for each query on ActivityNet Captions are displayed with green boxes representing the relevant video and red boxes for irrelevant videos.}
  \label{supp_vis}
\end{figure}

In order to validate the effectiveness of GMMFormer v2 in a qualitative manner, we present several typical examples from ActivityNet Captions in Fig. \ref{supp_vis}. Upon examining these retrieval results, it becomes apparent that our GMMFormer v2 model is capable of returning more precise retrieval outcomes compared to other competitive models, such as GMMFormer and MS-SL.

\section{Limitations and Future Work}

For a fair comparison, we follow prior arts~\cite{MSSL, DKD, GMMFORMER} to use pretrained models (\eg ResNet~\cite{he2016deep} and RoBERTa~\cite{roberta}) to extract features beforehand. This might lead to the unsatisfied retrieval performance and we will utilize more advanced multi-modal encoders (\eg CLIP~\cite{CLIP} and ViCLIP~\cite{wang2023internvid}) as our backbone and train our model in an end-to-end manner.

\section{Society Impacts}

Our solution of uncertainty-aware text-clip matching enhances the retrieval performance of partially relevant video retrieval models, making it easier for people to get the video content they want in their daily entertainment. However, PRVR models are still in development and may retrieve some video content that people do not want to see.

\end{document}